\documentclass[11pt,a4paper]{article}
\usepackage[hyperref]{acl2019}
\usepackage{times}
\usepackage{multirow}
\usepackage{latexsym}
\usepackage{placeins}
\usepackage{amsmath}
\usepackage{float}
\usepackage{url}
\usepackage{graphicx}
\usepackage{tablefootnote}
\graphicspath{{./image/}}
\aclfinalcopy % Uncomment this line for the final submission
\def\aclpaperid{1816} %  Enter the acl Paper ID here

\usepackage{amssymb}
\usepackage{autonum}

\newcommand\highres{\textsc{HighRES}}
\newcommand\ptgen{\textsc{PtGen}}

\newcommand\tconv{\textsc{TConvS2S}}
\newcommand\xsum{\textsc{XSum}}
\newcommand\hrouge{\textsc{HROUGE}}
\newcommand\rouge{\textsc{ROUGE}}

\definecolor{forestgreen}{HTML}{009B55}
\definecolor{sepia}{HTML}{671800}
\definecolor{midnightblue}{HTML}{006795}
\definecolor{orangered}{HTML}{ED135A}

\title{\highres: Highlight-based Reference-less Evaluation of Summarization}

\author{Hardy$^1$ \quad Shashi Narayan$^2$\thanks{\ The work was primarily done while Shashi was still at School of Informatics, University of Edinburgh.} \quad Andreas Vlachos$^{1,3}$ \\
  $^{1}$Department of Computer Science, University of Sheffield \quad $^{2}$Google Research\\
  $^{3}$Department of Computer Science and Technology, University of Cambridge \\
  \tt{\small \url{hhardy2@sheffield.ac.uk}}, \tt{\small \url{shashinarayan@google.com}}, \tt{\small \url{av308@cam.ac.uk}} \\}

\date{}

\begin{document}
\maketitle
\begin{abstract}
  There has been substantial progress in summarization research enabled by the availability of novel, often large-scale, datasets and recent advances on neural network-based approaches. 
  However, manual evaluation of the system generated summaries is inconsistent due to the difficulty the task poses to human non-expert readers.
  To address this issue, we propose a novel approach for manual evaluation,  \textsc{High}light-based \textsc{R}eference-less \textsc{E}valuation of \textsc{S}ummarization (\highres), in which summaries are assessed by multiple annotators against the source document via manually highlighted salient content in the latter. Thus summary assessment on the source document by human judges is facilitated,  while the highlights can be used for evaluating multiple systems.
   To validate our approach we employ crowd-workers to augment with highlights a recently proposed dataset and compare two state-of-the-art systems. We demonstrate that \highres{} improves inter-annotator agreement in comparison to using the source document directly, while they help emphasize differences among systems that would be ignored under other evaluation approaches.\footnote{Our dataset and code are available at \url{https://github.com/sheffieldnlp/highres}}
\end{abstract}

\section{Introduction}

Research in automatic summarization has made headway over the years with single document summarization as the front-runner due to the availability of large datasets \citep{Sandhaus2008,Hermann2015,narayan18xsum} which has enabled the development of novel methods, many of them employing recent advances in neural networks 
\citep[\textit{inter alia}]{See2017,Narayan2018,Pasunuru2018a}. 

\begin{figure}[t!]
    \centering
    \includegraphics[width=7.6cm]{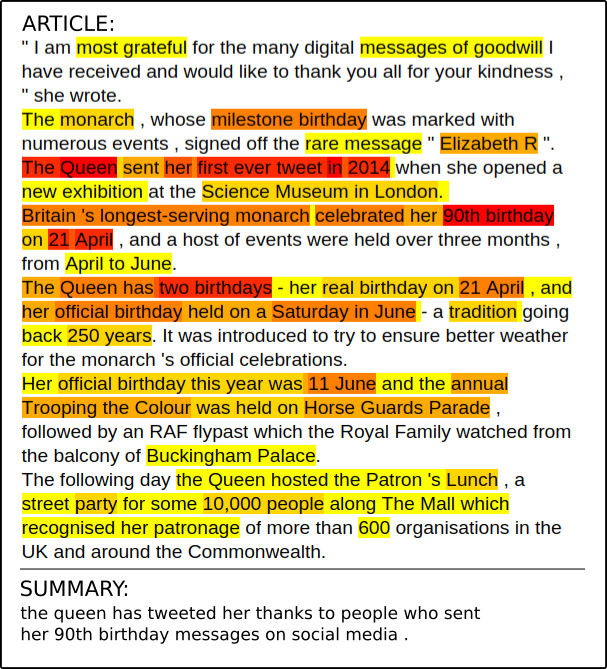}
    \caption{Highlight-based evaluation of a summary. Annotators to evaluate a summary (bottom) against the highlighted source document (top) presented with a heat map marking the salient content in the document; the darker the colour, the more annotators deemed the highlighted text salient.
    }
  \label{image:heatmap}
\end{figure}

Measuring progress in summarization is difficult, as the task has as input a source document consisting of multiple sentences and methods need to generate a shorter text that expresses the salient information of the source fluently and succinctly. Thus there can be multiple equally good summaries for the same source document as not all salient information can fit in a given summary length, while even extractive methods that select complete sentences are not guaranteed to produce a coherent summary overall.

The most consistently used evaluation approach is comparison of the summaries produces against reference summaries via 
automatic measures such as ROUGE \citep{Lin2004} and its variants. However, automatic measures are unlikely to be sufficient to measure performance in summarization \citep{schluter:2017:EACLshort}, also known for other tasks in which the goal is to generate natural language \citep{novikova2017we}. Furthermore, the datasets typically considered have a single reference summary, as obtaining multiple ones increases dataset creation cost, thus evaluation against them is likely to exhibit reference bias \citep{Louis2013,fomicheva2016reference}, penalizing summaries containing salient content different from the reference. 

For the above reasons manual evaluation is considered necessary for measuring progress in summarization. However, the intrinsic difficulty of the task has led to research without manual evaluation or only fluency being assessed manually.  Those that conduct manual assessment of the content, typically use a single reference summary, either directly \citep{Celikyilmaz2018, Tan2017} or through questions \citep{narayan18xsum,Narayan2018} and thus are also likely to exhibit reference bias.

In this paper we propose a novel approach for manual evaluation, \textsc{High}light-based \textsc{R}eference-less \textsc{E}valuation of document \textsc{S}ummarization (\highres), in which a summary is assessed against the source document via manually highlighted salient content in the latter (see Figure~\ref{image:heatmap} for an example). Our approach avoids reference bias, as the multiple highlights obtained help consider more content than what is contained in a single reference. The highlights are not dependent on the summaries being evaluated but only on the source documents, thus they are reusable across studies, and they can be crowd-sourced more effectively than actual summaries. 
Furthermore, we propose to evaluate the clarity of a summary separately from its fluency, as they are different dimensions.
Finally, \highres\ provides absolute instead of ranked evaluation, thus the assessment of a system can be conducted and interpreted without reference to other systems.

To validate our proposed approach we use the recently introduced e\textsc{X}treme \textsc{Sum}marization dataset \citep[\xsum,][]{narayan18xsum} to evaluate two state-of-the-art abstractive summarization methods, Pointer Generator Networks \citep{See2017} and Topic-aware Convolutional Networks \citep{narayan18xsum}, using crowd-sourcing for both highlight annotation and quality judgments.

We demonstrate that \highres{} improves inter-annotator agreement in comparison to using the source document directly, while they help emphasize differences among systems that would be ignored under other evaluation approaches, including reference-based evaluation. Furthermore, we show that the clarity metric from the DUC \citep{dang2005overview} must be measured separately from ``fluency'', as judgments for them had low correlation. %We demonstrate our findings with both quantitative and qualitative analyses.
Finally, we make the highlighted \xsum\ dataset, codebase to replicate the crowd-sourcing experiments and all other materials produced in our study publicly available.
\begin{table*}[t!]
\centering
\small
\begin{tabular}{r | c | c c c c c c c | c c | c c r}

\hline
\rotatebox[origin=c]{0}{\textbf{Systems}} & 
\rotatebox[origin=l]{90}{\textbf{No Manual Eval}} & 
\rotatebox[origin=l]{90}{\textbf{Pyramid}} & \rotatebox[origin=l]{90}{\textbf{QA}} & \rotatebox[origin=l]{90}{\textbf{Correctness}} &
\rotatebox[origin=l]{90}{\textbf{Fluency}} &
\rotatebox[origin=l]{90}{\textbf{Clarity}} & \rotatebox[origin=l]{90}{\textbf{Recall}} & \rotatebox[origin=l]{90}{\textbf{Precision}} & \rotatebox[origin=l]{90}{\textbf{Absolute}} & \rotatebox[origin=l]{90}{\textbf{Relative}} & \rotatebox[origin=l]{90}{\textbf{With Reference}} & \rotatebox[origin=l]{90}{\textbf{With Document}} & \rotatebox[origin=l]{90}{\textbf{With Ref. \& Doc.}} \\
\hline
\citet{See2017}        &  $\checkmark$   &   &   &    &   &  &  &   &    &      &     &    &   \\
\citet{Lin2018}     &  $\checkmark$   &   &    &   &   &  &   &    &      &     &    &     &  \\
\citet{Cohan2018a}     &  $\checkmark$   &   &    &   &  &  &   &    &      &     &    &     &  \\
\citet{Liao2018a}      &  $\checkmark$   &   &    &   &  &  &   &    &      &     &    &    &  \\
\citet{Kedzie2018}     &  $\checkmark$   &   &    &   &  &  &   &    &      &     &    &    & \\
\citet{Amplayo2018a}   &  $\checkmark$   &   &    &   &  &  &   &    &      &    &     &    & \\
\citet{Jadhav2018}    &  $\checkmark$   &   &    &   &  &  &   &    &      &    &     &   &  \\
\citet{li2018guiding}        &  $\checkmark$   &   &    &   &  &  &   &      &     &    &     &    & \\
\citet{Pasunuru2018a}  &  $\checkmark$   &   &    &   &  &  & &      &     &    &     &    & \\
\citet{Cao2018a}       &  $\checkmark$   &   &    &   &  &    &    &      &     &    &     &    & \\
\citet{Sakaue2018a}    &  $\checkmark$   &   &    &   &   &   &    &      &     &    &     &   & \\
\citet{Celikyilmaz2018}      &    &  &  &    &  &  &  $\checkmark$ &  $\checkmark$    &  $\checkmark$   &  $\checkmark$     &  $\checkmark$    &  $\checkmark$   & \\
\citet{Chen2018a}     &   &   &  & $\checkmark$   & & &  $\checkmark$ &  $\checkmark$    &    &  $\checkmark$     &     &    &  $\checkmark$   \\

\citet{Guo2018a}     &      &   &  &   &  $\checkmark$  & &  &  $\checkmark$    &    &  $\checkmark$     &     &    &  $\checkmark$  \\
\citet{Hardy2018}    &     &   &  &   & $\checkmark$  & &   &     & $\checkmark$  &      &     &    &   \\
\citet{Hsu2018}      &      &  &  &    &  $\checkmark$  & &  $\checkmark$ &  $\checkmark$    &  $\checkmark$   &      &     &  $\checkmark$   &  \\
\citet{Krishna2018a}    &      &   &  &   &  &  &  $\checkmark$ &     &    &  $\checkmark$     &     &  $\checkmark$   &    \\
\citet{Kryscinski2018}    &    &   &  &   &  $\checkmark$ & &  $\checkmark$ & &  $\checkmark$   &      &  & $\checkmark$ &   \\
\citet{Li2018a}     &       &   &   & $\checkmark$ &  &  &   &     &  $\checkmark$   &      &     &     &    \\
\citet{narayan-sidenet18}    &     & &   &    &  $\checkmark$ & &  &     &    & $\checkmark$ &  &  $\checkmark$   &    \\
\citet{narayan18xsum}       &      &   &  $\checkmark$ &    &  $\checkmark$ &  &  $\checkmark$ &     &  $\checkmark$  &  $\checkmark$     &   $\checkmark$  &  $\checkmark$   &   \\
\citet{Narayan2018}     &     &   & $\checkmark$ &   &  $\checkmark$  & &  $\checkmark$ &     &  $\checkmark$  &  $\checkmark$   &  $\checkmark$   &  $\checkmark$  &   \\
\citet{Peyrard2018a}    &    &   &  &   &  &  &  $\checkmark$ &  $\checkmark$    &  $\checkmark$   &      &  $\checkmark$    &    &   \\
\citet{ShafieiBavani2018}   &      &  $\checkmark$ &  &    &  &  &  &     &    &      &     &    &  \\
\citet{Song2018}      &    &   &  & $\checkmark$   &  $\checkmark$ &  &  $\checkmark$ &     &  $\checkmark$   &      &     &  $\checkmark$   & \\
\citet{Yang2017b}      &    &  &  &    &  &  &  $\checkmark$ &     &    &  $\checkmark$     &  $\checkmark$    &    & \\
\highres\ (ours)    &   &   &  &   & $\checkmark$ & $\checkmark$ & $\checkmark$  &  $\checkmark$   & $\checkmark$  &      &     &  $\checkmark$  &   \\
\hline
\end{tabular}
\caption{Overview of manual evaluations conducted in recent summarization systems. We categorize them in four dimensions: the first columns presents papers that do not report on human evaluation; the second column identifies matrices used for evaluating content (``\textit{Pyramid}'', ``\textit{QA}'', ``\textit{Correctness}'', ``\textit{Recall}'' and ``\textit{Precision}'') and quality (``\textit{Clarity}'', ``\textit{Fluency}'') of summaries; the third column focuses if the system ranking reported by humans on content evaluation were ``\textit{Absolute}'' or ``\textit{Relative}''; and finally, the fourth column evaluates if summaries were evaluated against the input document (``\textit{With Document}''), the reference summary (``\textit{With Reference}'') or both (``\textit{With Ref. \& Doc.}'').}
\label{tab:litreview}
\end{table*}

\section{Literature Review}
\label{sec:review}

In recent years, summarization literature has investigated different means of conducting manual evaluation. We study a sample of 26 recent papers from major ACL conferences and outline the trends of manual evaluation in summarization in Table~\ref{tab:litreview}. From 26 papers, 11 papers \citep[e.g.,][]{See2017,Kedzie2018,Cao2018a} did not conduct any manual evaluation.
% , they are not shown in Table~\ref{tab:litreview}. 
Following the Document Understanding Conference \citep[DUC, ][]{dang2005overview}, a majority of work has focused on evaluating the content and the linguistic quality of summaries \cite{Nenkova:2005:ATS}. However, there seems to be a lack of consensus on how a summary should be evaluated: (i) Should it be evaluated relative to other summaries or standalone in absolute terms? and (ii) What would be a good source of comparison: the input document or the reference summary? The disagreements on these issues result in authors evaluating their summaries often (11 out of 26 papers) using automatic measures such as ROUGE \cite{Lin2004} despite of its limitations \cite{schluter:2017:EACLshort}. 
In what follows, we discuss previously proposed %manual evaluation
approaches along three axes: evaluation metrics, relative vs. absolute, and the choice of reference.

\paragraph{Evaluation Metrics}
Despite differences in the exact definitions,
the majority \citep[e.g.,][]{Hsu2018,Celikyilmaz2018,narayan18xsum,Chen2018a,Peyrard2018a} agree on both or either one of two broad quality definitions: {\em coverage} determines how much of the salient content of the source document is captured in the summary, and {\em informativeness}, how much of the content captured in the summary is salient with regards to the original document. These measures correspond to ``\textit{recall}'' and ``\textit{precision}'' metrics respectively in Table~\ref{tab:litreview}, notions that are commonly used in information retrieval and information extraction literature. \citet{Clarke2010} proposed a question-answering based approach to improve the agreement among human evaluations for the quality of summary content, which was recently employed by \citet{narayan18xsum} and \citet{Narayan2018} (QA in Table~\ref{tab:litreview}). In this approach,  questions were created first from the reference summary and then the system summaries were judged with regards to whether they enabled humans to answer those questions correctly. \citet{ShafieiBavani2018}, on the other hand, used the ``Pyramid'' method \citep{Nenkova2004a} which requires summaries to be annotated by experts for salient information. 
A similar evaluation approach is the factoids analysis by  \citet{teufel2004evaluating} which evaluates the system summary against factoids, a representation based on atomic units of information, that are extracted from multiple gold summaries. However, as in the case of the ``Pyramid'' method, extracting factoids  %would 
requires experts annotators. 
%and are laborious as it would necessary to create many gold summaries from a single document to gather sufficient factoids. 
%Our approach adopts a principle similar to factoid analysis but use the source article directly for the highlight base and does not require expert annotators.
Finally, a small number of work evaluates the "Correctness" \citep{Chen2018a,Li2018a,Chen2018a} of the summary, similar to fact checking \cite{vlachos-riedel:2014:W14-25}, which can be a challenging task in its own right.

The linguistic quality of a summary encompasses many different qualities such as fluency, grammatically, readability, formatting, naturalness and coherence. Most recent work uses 
a single human judgment
to capture all linguistic qualities of the summary \cite{Hsu2018,Kryscinski2018,narayan18xsum,Song2018,Guo2018a}; we group them under ``Fluency'' in Table~\ref{tab:litreview} with an exception of ``Clarity'' which was evaluated in the DUC evaluation campaigns \citep{dang2005overview}. The ``Clarity'' metric puts emphasis in easy identification of noun and pronoun phrases in the summary which is a different dimension than ``Fluency'', as a summary may be fluent but difficult to be understood due to poor clarity.

\paragraph{Absolute vs Relative Summary Ranking.} 
In relative assessment of summarization, 
annotators are shown two or more summaries and are asked to rank them according to the dimension at question \citep{Yang2017b,Chen2018a,narayan-sidenet18,Guo2018a,Krishna2018a}. The relative assessment is often done using the paired comparison \citep{Thurstone1994} or the best-worst scaling \citep{Woodworth1991,Louviere2015}, to improve inter-annotator agreement. On the other hand,
absolute assessment of summarization \cite{Li2018a,Song2018,Kryscinski2018,Hsu2018,Hardy2018} 
is often done using the Likert rating scale \citep{Likert1932} where a summary is assessed on a numerical scale. 
Absolute assessment was also employed in combination with the question answering approach for content evaluation
\citep{narayan18xsum,mendesetalnaacl19}. Both approaches, relative ranking and absolute assessment, have been investigated extensively in Machine Translation \citep{Bojar2016, Bojar2017}.  
 Absolute assessment correlates highly with the relative assessment without the bias introduced by having a simultaneous assessment of several models \citep{Bojar2011}.

\paragraph{Choice of Reference.} 
The most convenient
way to evaluate a system summary is to assess it against the reference summary \citep{Celikyilmaz2018,Yang2017b,Peyrard2018a}, as this typically requires less effort than reading the source document. The question answering approach of \citet{narayan18xsum,Narayan2018} also falls in this category, as the questions were written using the reference summary. However, summarization datasets are limited to a single reference summary per document \citep{Sandhaus2008,Hermann2015,newsroom_N181065,narayan18xsum} thus evaluations using them is prone to reference bias \citep{Louis2013}, also a known issue in machine translation evaluation \citep{fomicheva2016reference}. A circumvention for this issue is to evaluate it against the source document \citep{Song2018,narayan-sidenet18,Hsu2018,Kryscinski2018}, asking judges to assess the summary after reading the source document. However this requires more effort and is known to lead to low inter-annotator agreement \citep{Nenkova2004a}. 

\section{\highres}
\label{sec:highres}

Our novel highlight-based reference-less evaluation does not suffer from reference bias as a summary is assessed against the source document with manually highlighted salient content. These highlights are crowd-sourced effectively without the need of expert annotators as required by the Pyramid method \citep{Nenkova2004a} or to generate reference summaries. Our approach improves over the ``Correctness'' or ``Fluency'' only measure for summarization by taking salience into account. Finally, the assessment of summaries against the document with  highlighted pertinent content facilitates an absolute evaluation of summaries with high inter-annotator agreement. 

Our evaluation framework comprises three main components: document highlight annotation, highlight-based content evaluation, and 
clarity and fluency evaluation. The second component, which evaluates the notions of ``Precision'' and ``Recall'' requires the highlights from the first one to be conducted. However, the highlight annotation needs to happen only once per document, and it can be reused to evaluate many system summaries, unlike the Pyramid approach \citep{Nenkova2004a} that requires additional expert annotation for every system summary being evaluated. The third component is independent of the others and can be run in isolation. In all components we employ crowd-workers as human judges, and implement appropriate sanity checking mechanisms to ensure good quality judgements. Finally, we present an extended version of ROUGE \cite{Lin2004} that utilizes the highlights to evaluate system summaries against the document; this demonstrates another use of the highlights for summarization evaluation. 

\subsection{Highlight Annotation}
\label{subsec:hannot}
In this part, we ask human judges to read the source document and then highlight words or phrases that are considered salient.
Each judge is allowed to highlight parts of the text at any granularity, from single words to complete sentences or even paragraphs. However we enforce a limit in the number of words to $\mathcal{K}$ that can be highlighted in total by a judge in a document, corresponding to the length of the summary expected. By employing multiple judges per document who are restricted in the amount of text that can be highlighted  we expect to have a more diverse and focused highlight from multiple judges which cover different viewpoints of the article. To ensure that each highlight is reliable, we performed a sanity check at the end of the task where we ask the judges to answer a True/False question based on the article. We rejected all annotations that failed to correctly answer the sanity check question.

\subsection{Highlight-based Content Evaluation}
\begin{figure*}[t!]
    \centering
    \fbox{\includegraphics[width=15.5cm]{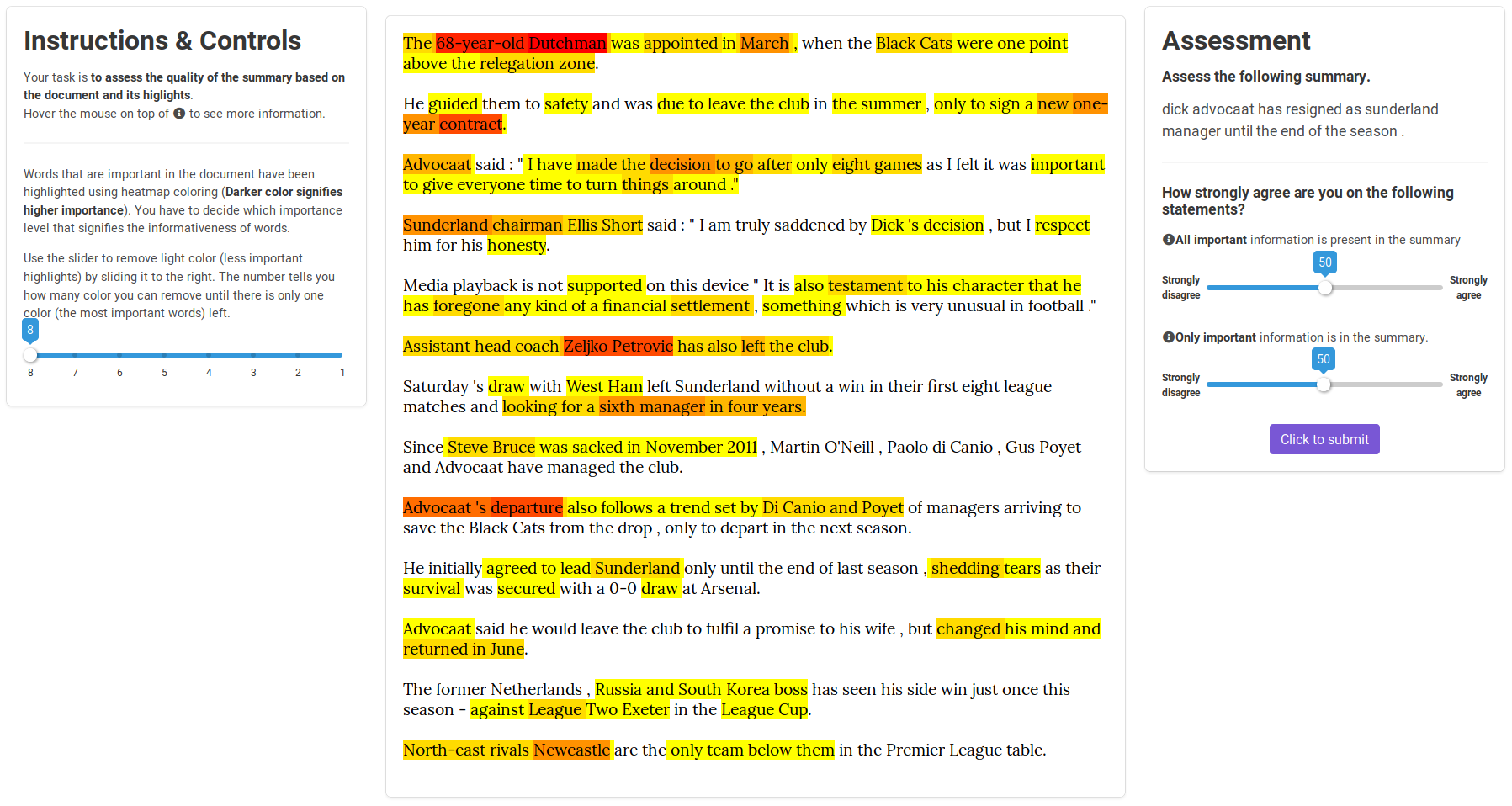}}
    \caption{The UI for content evaluation with highlight. Judges are given an article with important words highlighted using heat map. Judges can also remove less important highlight color by sliding the scroller at the left of the page. At the right of the page judges give the recall and precision assessment by sliding the scroller from 1 to 100 based on the given summary quality.}
    \label{image:highlightbasedevaluation}
\end{figure*}

In this component, we present human judges a document that has been highlighted using heatmap coloring and a summary to assess.
% (see Figure~\ref{image:heatmap} for an example). 
We ask our judges to assess the summary for (i) \textit{`All important information is present in the summary'} and (ii) \textit{`Only important information is in the summary.'} The first one is the recall (content coverage) measure and the second, the precision (informativeness) measure. All the ratings were collected on a 1-100 Likert scale \citep{Likert1932}. Figure~\ref{image:highlightbasedevaluation} shows the content evaluation user interface where salient parts of the document are highlighted.
As with the highlight annotation, %to ensure that each rating is reliable, 
we performed the same form of sanity check to the one in the highlight annotation task.

\subsection{Clarity and Fluency Evaluation}
In this part, we give the judges only the summary and ask them to rate it on clarity and fluency. For {\em clarity}, each judge is asked whether the summary is easy to be understood, i.e.\ 
there should be no difficulties in identifying the referents of the noun phrases (every noun/place/event should be well-specified) or understanding the meaning of the sentence. For {\em fluency}, each judge is asked whether the summary sounds natural and has no grammatical problems. % that makes the text difficult to read.
While fluency is often evaluated in recent work, clarity, while first introduced in DUC evaluations, has recently been ignored in manual evaluation, despite that it captures a different dimension of summarization quality.

To ensure that the judgments for clarity and fluency are not affected by each other (poor fluency can affect clarity, but a summary can have perfect fluency but low clarity), we evaluate each metric separately. We ask the judges to evaluate multiple summaries per task with each dimension in its own screen. For sanity checking, we insert three artificial summaries of different quality (good, mediocre and bad summaries). The good summary is the unedited one, while the others are generated from sentences randomly sampled from the source document. For the mediocre summary, some words are edited to introduce some grammatical or syntactic errors while for the bad summary, the words are further scrambled. We reject judgements that failed to pass this criteria: bad $<$ mediocre $<$ good.

\subsection{Highlight-based ROUGE Evaluation} 
\label{subsec:hrouge}

Our Highlight-based \rouge{} (we refer to it as \hrouge{}) formulation is similar to the original ROUGE with the difference that the n-grams are weighted by the number of times they were highlighted. One benefit of \hrouge{} is that it introduces saliency into the calculation without being reference-based as in \rouge{}. Implicitly \hrouge{} considers multiple summaries as the highlights are obtained from multiple workers. 

Given a document $\mathcal{D}$ as a sequence of $m$ tokens $\{w_1, \ldots, w_m\}$, annotated with $\mathcal{N}$ highlights, we define the weight $\beta_g^n \in [0,1]$ for an $n$-gram $g$ as: 
\begin{equation}
    \beta_g^n = \frac{\displaystyle\sum_{i=1}^{m-(n-1)} \Bigg[\frac{\sum_{j=i}^{i+n-1} \frac{\mathrm{NumH}(w_j)}{\mathcal{N}}}{n}\Bigg]_{w_{i:i+n-1} == g}}{\displaystyle\sum_{i=1}^{m-(n-1)} [1]_{w_{i:i+n-1} == g} }
\end{equation}
\noindent where, $[x]_y$ is an indicator function which returns $x$ if $y$ is true and $0$, otherwise.  $\mathrm{NumH}(w_j) = \sum_{k=1}^{\mathcal{N}} \frac{\mathrm{len}(H_k)}{\mathcal{K}} [1]_{w_j \in H_k}$ is a function which returns the number of times word $w_j$ is highlighted by the annotators out of $\mathcal{N}$ times weighted by the lengths of their highlights; $H_k$ is the highlighted text by the $k$-th annotator and  $\mathcal{K}$ is the maximum allowed length of the highlighted text (see Section~\ref{subsec:hannot}). $\mathrm{NumH}(w_j)$ gives less importance to  annotators with highlights with few words. In principle, if an $n$-gram is highlighted by every crowd-worker and the length of the highlight of each crowd-worker is $\mathcal{K}$, the $n$-gram $g$ will have a maximum weight of $\beta_g^n = 1$.

The HROUGE scores for a summary $\mathcal{S}$ can then be defined as:
\begin{align}
    \text{HR}_{\mathrm{rec}}^n &= \frac{\displaystyle\sum_{g \in n\operatorname{-gram}(\mathcal{S})} \beta_g^n\, \text{count}(g, \mathcal{D} \cap \mathcal{S})}{\displaystyle\sum_{g \in n\operatorname{-gram}(\mathcal{D})} \beta_g^n\,\text{count}(g, \mathcal{D})} \\
    \text{HR}_{\mathrm{pre}}^n &= \frac{\displaystyle\sum_{g \in n\operatorname{-gram}(\mathcal{S})} \beta_g^n\, \text{count}(g, \mathcal{D} \cap \mathcal{S})}{\displaystyle\sum_{g \in n\operatorname{-gram}(\mathcal{S})} \text{count}(g, \mathcal{S})}
\end{align}
\noindent $\text{HR}_{\mathrm{rec}}^n$ and $\text{HR}_{\mathrm{pre}}^n$ are the HROUGE recall and precision scores; $\text{count}(g, \mathcal{X})$ is the maximum number of $n$-gram $g$ occurring in the text $\mathcal{X}$. The weight in the denominator of $\text{HR}_{\mathrm{pre}}^n$ is uniform ($\beta^n_g = 1$) for all $g$ because if we weighted according to the highlights, words in the summary that are not highlighted in the original document would be ignored. This would result in $\text{HR}_{\mathrm{pre}}^n$ not penalizing summaries for containing words that are likely to be irrelevant as they do not appear in the highlights of the document. It is important to note \hrouge{} has an important limitation in that it penalizes abstractive summaries that do not reuse words from the original document. This is similar to \rouge{} penalizing summaries for not reusing words from the reference summaries, however the highlights allow us to implicitly consider multiple references without having to actually obtain them.  %produced with lower annotation effort.

\section{Summarization Dataset and Models}
\label{sec:data-models}

We use the extreme summarization dataset \citep[\xsum,][]{narayan18xsum}\footnote{\url{https://github.com/EdinburghNLP/XSum}} which comprises BBC articles paired with their single-sentence summaries, provided by the journalists writing the articles. The summary in the \xsum\ dataset demonstrates a larger number of novel $n$-grams compared to other popular datasets such as CNN/DailyMail \citep{Hermann2015} or NY Times \citep{Sandhaus2008} as such it is suitable to be used for our experiment since the more abstractive nature of the summary renders automatic methods such as ROUGE less accurate as they rely on string matching, and thus calls for human evaluation for more accurate system comparisons. Following \citet{narayan18xsum}, we didn't use the whole test set portion, but sampled 50 articles from it for our highlight-based evaluation.  

We assessed summaries from two state-of-the-art abstractive summarization systems using our highlight-based evaluation: (i) the Pointer-Generator model (\ptgen) introduced by \citet{See2017} is an RNN-based abstractive systems which allows to copy words from the source text, and (ii) the Topic-aware Convolutional Sequence to Sequence model (\tconv) introduced by \citet{narayan18xsum} is an   abstractive model which is conditioned on the article's topics and based entirely on Convolutional Neural Networks.
We used the pre-trained models\footnote{Both models were trained using the standard cross-entropy loss to maximize the likelihood of the reference summary given the document.} provided by the authors to obtain summaries from both systems for the documents in our test set.

\section{Experiments and Results}
\label{sec:exp-res}

All of our experiments are done using the Amazon Mechanical Turk platform.We develop three types of Human Intelligence Tasks (HITs): highlight annotation, highlight-based content evaluation, and fluency and clarity evaluation. In addition, we elicited human judgments for content evaluation in two more ways: we assessed system summaries against the original document (without highlights) and against the reference summary. The latter two experiments are intended as the comparison for our proposed highlight-based content evaluation.

\subsection{Highlight Annotation} 

We collected highlight annotations from 10 different participants for each of 50 articles. For each annotation, we set  $\mathcal{K}$, the maximum number of words to highlight, to 30. Our choice reflects the average length (24 words) of reference summaries in the \xsum\ dataset. To facilitate the annotation of BBC news articles with highlights, we asked our participants to adapt the 5W1H (Who, What, When, Where, Why and How) principle \citep{Robertson1946} that is a common practice in journalism. The participants however were not obliged to follow this principle and were free to highlight content as they deem fit. 

The resulting annotation exhibits a substantial amount of variance, confirming the intuition that different participants are not expected to agree entirely on what is salient in a document. 
On average, the union of the highlights from 10 annotators covered 38.21\% per article and  33.77\% of the highlights occurred at the second half of the article. This shows that the judges did not focus only on the beginning of the documents but annotated all across the document.

Using Fleiss Kappa \citep{Josep1971} on the binary labels provided by each judge on each word (highlighted or not) we obtained an average agreement of 0.19 for the 50 articles considered. The low agreement score does not indicate a poor annotation process necessarily; we argue that this is primarily due to the annotators having different opinions on which parts of an article are salient. The article with the highest agreement (0.32) has more focused highlights, whereas the article with the lowest agreement (0.04) has highlights spread all over (both articles can be seen in the supplementary materials). Interestingly, the reference summary on the highest agreement article appears to be more informative of its content when the annotator agreement is high; the reference summary on the lowest agreement article is more indicative, i.e., it does not contain any informative content from the article but only to inform the reader about the article's topic and scope. These results confirm that the annotation behaviour originates from the nature of the document and the summary it requires, and validates our highlight annotation setup.

\begin{table}[t!]
\small
\begin{tabular}{l|p{0.4cm}p{0.6cm}|p{0.4cm}p{0.4cm}|p{0.4cm}p{0.4cm}}
\hline
\multirow{3}{*}{\textbf{Model}}                & \multicolumn{2}{c|}{\textbf{Highlight}}                              & \multicolumn{2}{c|}{\textbf{Non High-}} & 
\multicolumn{2}{c}{\textbf{Reference}}\\
\multirow{2}{*}{}                
& \multicolumn{2}{c|}{\textbf{-based}}                              & \multicolumn{2}{c|}{\textbf{light-based}} & 
\multicolumn{2}{c}{\textbf{-based}}\\
& \textbf{Prec} & \textbf{Rec} 
& \textbf{Prec} & \textbf{Rec}
& \textbf{Prec} & \textbf{Rec} \\ \hline
\tconv{} & 57.42 & 49.95 & 52.55 & 41.04 & 46.75 & 36.45             \\
\ptgen{} & 50.94 & 44.41 & 48.57 & 39.21  & 44.24 & 38.24             \\
Reference & 67.90 & 56.83 & 66.01 & 52.45  & --- & --- \\
\hline
\end{tabular}
\caption{Results of content evaluation of summaries against documents with highlights, documents without highlights and reference summaries.}
\label{table:summresult}
\end{table}

\begin{table}[t!]
\small
\begin{tabular}{l|cc|cc}
\hline
\multirow{2}{*}{\textbf{Model}} & 
\multicolumn{2}{c|}{\textbf{Highlight-based}} & 
\multicolumn{2}{c}{\textbf{Non Highlight-based}} \\
&\textbf{Prec} & \textbf{Rec} & 
 \textbf{Prec} & \textbf{Rec} \\ \hline
\tconv{} & 0.67 & 0.80 & 0.75                             & 0.83  \\
\ptgen{} & 0.73 & 0.86 & 0.73   & 0.90  \\
Reference                             & 0.49                           & 0.63 & 0.48   & 0.67 \\
\hline

\end{tabular}
\caption{Coefficient of variation (lower is better) for evaluating summaries against documents with and without highlights.}
\label{table:summresultcv}
\end{table}

\subsection{Content Evaluation of Summaries} 
\label{subsec:conteval}

We assessed the summaries against (i) documents with highlights (Highlight-based), (ii) original documents without highlights (Non Highlight-based) and (iii) reference summaries (Reference-based). For each setup, we collected judgments from 3 different participants for each model summary. Table \ref{table:summresult} and \ref{table:summresultcv} presents our results.

Both the highlight-based and non-highlight based assessment of summaries agree on the ranking among \tconv, \ptgen\ and Reference. Perhaps unsurprisingly human-authored summaries were considered best, whereas, \tconv\ was ranked 2nd, followed by \ptgen. However, the performance difference in \tconv\ and \ptgen\ is greatly amplified when they are evaluated against document with highlights (6.48 and 5.54 Precision and Recall points) compared to when evaluated against the original documents (3.98 and 1.83 Precision and Recall points). The performance difference is lowest when they are evaluated against the reference summary (2.51 and -1.79 Precision and Recall points). The superiority of \tconv\ is expected; \tconv\ is better than \ptgen\ for recognizing pertinent content and generating informative summaries due to its ability to represent high-level document knowledge in terms of topics and long-range dependencies \citep{narayan18xsum}.

We further measured the agreement among the judges using the coefficient of variation \citep{everitt2006cambridge} from the aggregated results. It is defined as the ratio between the sample standard deviation and sample mean. It is a scale-free metric, i.e.\ its results are comparable across measurements of different magnitude. Since, our sample size is  small (3 judgements per summary), we  use the unbiased version \citep{sokal1995biometry} as $cv = (1 + \frac{1}{4n})\frac{\sigma}{\bar{x}}$, where $\sigma$ is the standard deviation, $n$ is the number of sample, and $\bar{x}$ is the mean.

We found that the highlight-based assessment in general has lower variation among judges than the non-highlight based or reference-based assessment. The assessment of \tconv\ summaries achieves 0.67 and 0.80 of Precision and Recall $cv$ points which are 0.08 and 0.03 points below when they are assessed against documents with no highlights, respectively. We see a similar pattern in Recall on the assessment of \ptgen\ summaries. Our results demonstrate that the highlight-based assessment of abstractive systems improve agreement among judges compared to when they are assessed against the documents without highlights or the reference summaries. The assessment of human-authored summaries does not seem to follow this trend, we report a mixed results (0.49 vs 0.48 for precision and 0.63 vs 0.67 for recall) when they are evaluated with and without the highlights.

\begin{table}[t!]
\small
\centering
\begin{tabular}{l | cr}
\hline
\textbf{Model} & \textbf{Fluency} & \textbf{Clarity} \\
\hline
\tconv\  & 69.51        & 67.19        \\
\ptgen\      & 55.24        & 52.49        \\
Reference      & 77.03        & 75.83       \\
\hline
\end{tabular}
\caption{Mean "Fluency" and "Clarity" scores for \tconv\ , \ptgen\   and Reference summaries. All the ratings were collected on a 1-100 Likert scale.}
\label{table:fluencyclarityresult}
\end{table}

\subsection{Clarity and Fluency Evaluation} 

Table \ref{table:fluencyclarityresult} shows the results of our 
fluency and clarity evaluations. Similar to our highlight-based content evaluation, human-authored summaries were considered best, whereas \tconv\  was  ranked  2nd  followed  by  \ptgen, on both measures. The Pearson correlation between fluency and clarity evaluation is 0.68 which shows a weak correlation; it confirms our hypothesis that the "clarity" captures different aspects from "fluency" and they should not be combined as it is commonly done. 

\begin{table}[t!]
\centering
\small
\begin{tabular}{l|cc|cc}
\hline
\multirow{2}{*}{\textbf{Model}} & \multicolumn{2}{c|}{\textbf{Unigram}} & \multicolumn{2}{c}{\textbf{Bigram}} \\
& Prec     & Rec  & Prec     & Rec     \\
\hline 
\multicolumn{5}{c}{\rouge\ (Original document)} \\
\hline
\tconv{}              & \textbf{77.17}    & 4.20    & 26.12    & 1.21    \\
\ptgen{}                  & 77.09    & \textbf{4.99}       & \textbf{28.75}    & \textbf{1.64}      \\
Reference              & 73.65    & 4.42    & 22.42    & 1.17      \\
\hline
\multicolumn{5}{c}{\hrouge\ (Highlights from the document)} \\
\hline
\tconv{}               &   \textbf{7.94}  &  5.42  & 3.30 &  2.11   \\
\ptgen{}                 & 7.90    & \textbf{6.46}      & \textbf{3.37}    & \textbf{2.64}     \\
Reference              &  7.31  & 5.73   &  2.39  & 1.84   \\
\hline
\end{tabular}
\caption{\hrouge-1 (unigram) and \hrouge-2 (bigram) precision, and recall scores for \tconv\ , \ptgen\  and Reference summaries.}
\label{table:rougeandhrougeresult}
\end{table}

\subsection{Highlight-based \rouge{} Evaluation} 

Table~\ref{table:rougeandhrougeresult} presents our \hrouge\ results assessing \tconv\ , \ptgen\  and Reference summaries with the highlights. To compare, we also report \rouge\ results assessing these summaries against the original document without highlights. In the latter case, \hrouge\ becomes the standard \rouge\ metric with $\beta^n_g=1$ for all $n$-grams $g$. 

Both \rouge{} and \hrouge{} favour method of copying content from the original document and penalizes abstractive methods, thus it is not surprising that \ptgen{} is superior to \tconv{}, as the former has an explicit copy mechanism. The fact that \ptgen{} is better in terms of \hrouge{} is also an evidence that the copying done by \ptgen{} selects salient content, thus confirming that the copying mechanism works as intended. 
When comparing the reference summaries against the original documents, both \rouge{} and \hrouge{} confirm that the reference summaries are rather abstractive as reported by \citet{narayan18xsum}, and they in fact score below the system summaries. Recall scores are very low in all cases which is expected, since the 10 highlights obtained per document or the documents themselves, taken together, are much longer than any of the summaries.

\section{Qualitative Analysis}
\label{sec:qanalysis}

\begin{figure}[t!]
    \centering
    \includegraphics[width=7.6cm]{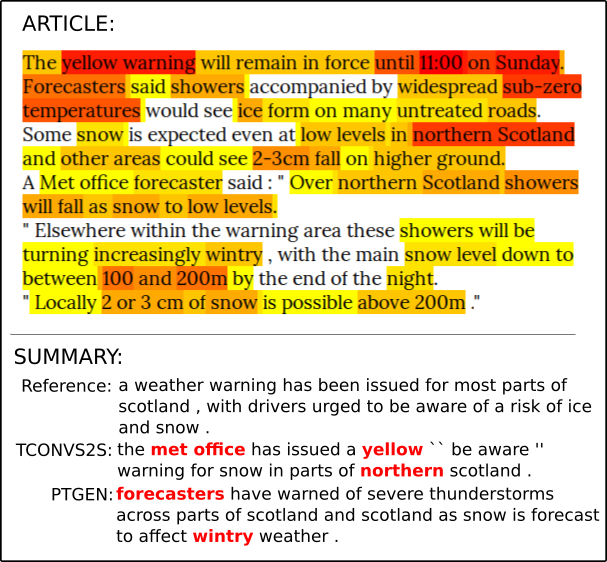}
    \caption{Highlighted article, reference summary, and summaries generated by \tconv\ and \ptgen. Words in red in the system summaries are highlighted in the article but do not appear in the reference.}
  \label{image:heatmap_small_2}
\end{figure}

\paragraph{\highres{} eliminates reference bias.}
The example presented in Figure \ref{image:heatmap_small_2} demonstrates how our highlight-based evaluation eliminates reference bias in summarization evaluation. The summaries generated by \tconv\ and \ptgen\ are able to capture the essence of the document, however, there are phrases in these summaries that do not occur in the reference summary. A reference-based evaluation would fail to give a reasonable score to these system summaries. The \highres{} however, would enable the judges to better evaluate the summaries without any reference bias.

\paragraph{Fluency vs Clarity.} Example in Table~\ref{table:fluencyclarityexample} shows disagreements between fluency and clarity scores for different summaries of the same article. From the example, we can see that the \tconv{} summary is fluent but is not easily understood in the context of `the duration of resignation', while the \ptgen{} summary has word duplication which lower the fluency and also lacking clarity due to several unclear words. 
\begin{table}[t!]
\small
\begin{tabular}{p{1.5cm}|p{3.1cm}|p{0.7cm}p{0.7cm}}
\hline
\textbf{Model}     & \textbf{Summary Text}  & \textbf{Fluency} & \textbf{Clarity} \\
\hline
\tconv{}  & dick advocaat has resigned as sunderland manager \textit{until the end of the season} .                                          & 92.80   & 44.33   \\
\ptgen{}     & sunderland have appointed \textit{former sunderland boss} dick advocaat as manager \textit{at the end of the season} to sign a \textit{new deal} . & 41.33   & 6.00    \\
\hline
\end{tabular}
\caption{\tconv{} and \ptgen{} showing a disagreement between fluency and clarity scores. We italicized words that are not clear in the summaries.}
\label{table:fluencyclarityexample}
\end{table}

\section{Conclusion and Future Work}
In this paper we introduced the \textsc{High}light-based \textsc{R}eference-less \textsc{E}valuation \textsc{S}ummarization (\highres) framework for manual evaluation.
The proposed framework avoids reference  bias
and provides absolute instead of ranked evaluation of the systems. Our experiments show that \highres{}  lowers the variability of the judges' content assessment, while helping expose the differences between systems. We also showed that by evaluating clarity we are able to capture a different dimension of summarization quality that is not captured by the commonly used fluency. We believe that our highlight-based evaluation is an ideal setup of abstractive summarization for three reasons: (i) highlights can be crowd sourced effectively without expert annotations, (ii) it avoids reference bias and (iii) it is not limited by n-gram overlap.
In future work, we would like to extend our framework to other variants of summarization e.g.\  multi-document. Also, we will explore ways of automating parts of the process, e.g.\ the highlight annotation. Finally, the highlights could also be used as training signal, as it offers content saliency information at a finer level than the single reference typically used.

\section*{Acknowledgments}

Hardy would like to thank the Indonesian government that
has sponsored his studies through
the Indonesia Endowment Fund for Education
(LPDP). Shashi Narayan and Andreas Vlachos were supported by the
EU H2020 SUMMA project (grant agreement
number 688139). The latter is also supported by the EPSRC grant eNeMILP
(EP/R021643/1).

\bibliography{acl2019_bib}
\bibliographystyle{acl_natbib}

\onecolumn
\newpage
\setcounter{section}{0}
\setcounter{figure}{0}
\setcounter{table}{0}
\section*{Supplementary Material}
\section{Highest and lowest highlight annotation agreement articles}
\begin{figure*}[ht!]
    \centering
    \begin{minipage}{.45\textwidth}
    \centering
    \includegraphics[height=7cm]{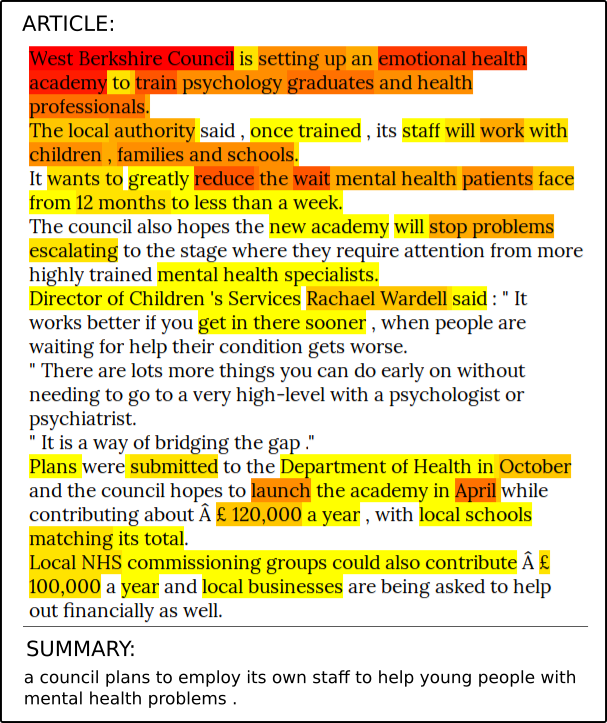}
    \end{minipage}
    \begin{minipage}{.45\textwidth}
    \centering
    \includegraphics[height=6.5cm]{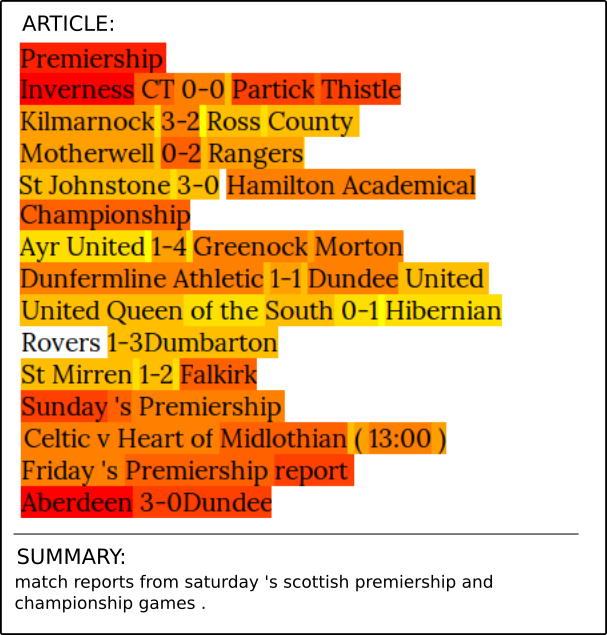}
    \end{minipage}
    \caption{Highlight annotation for the documents with the highest (left) and lowest (right) agreement. We also show their reference summaries at the bottom.}
  \label{image:heatmap_both}
\end{figure*}
\newpage
\section{HighRES User Interface Screenshots}
\begin{figure}[ht!]
    \centering
    \fbox{\includegraphics[width=17cm]{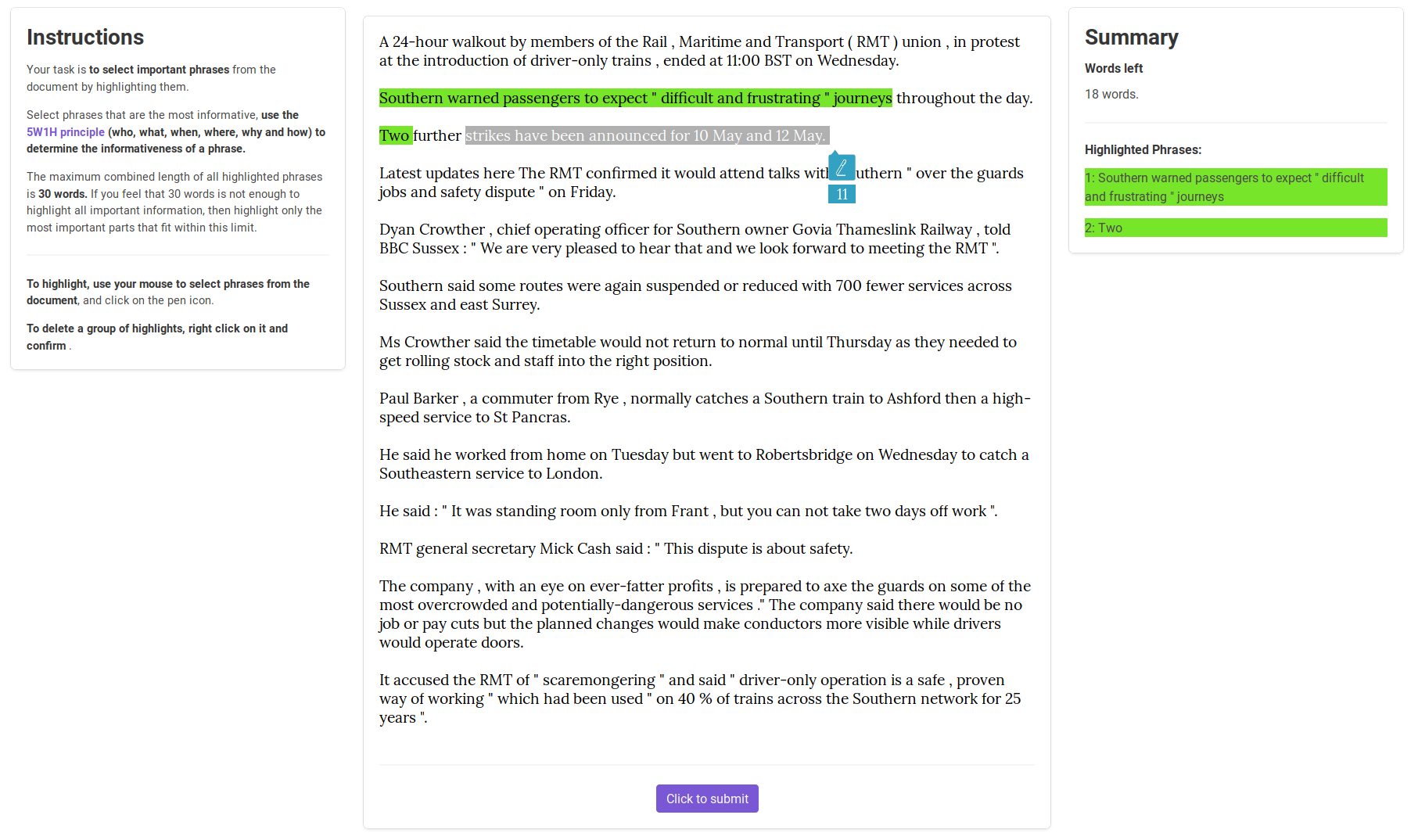}}
    \caption{The UI for highlight annotation. Judges are given an article and asked to highlight words or phrases that are important in the article.}
\end{figure}
\begin{figure}[ht!]
    \centering
    \fbox{\includegraphics[width=17cm]{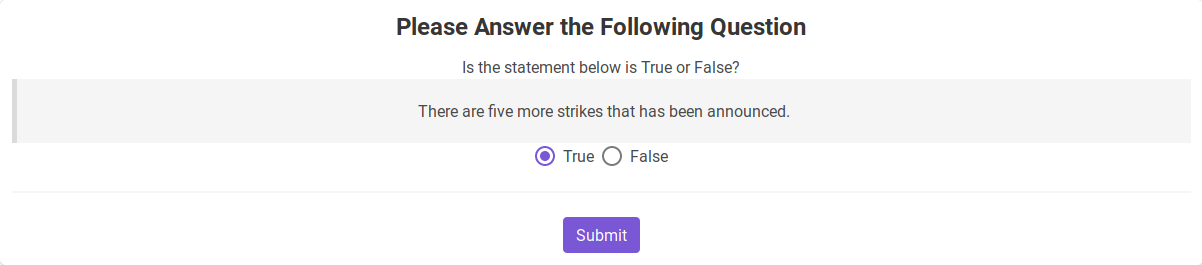}}
    \caption{The sanity checking question at the end of the annotation task.}
\end{figure}

\begin{figure}[ht!]
    \centering
    \fbox{\includegraphics[width=17cm]{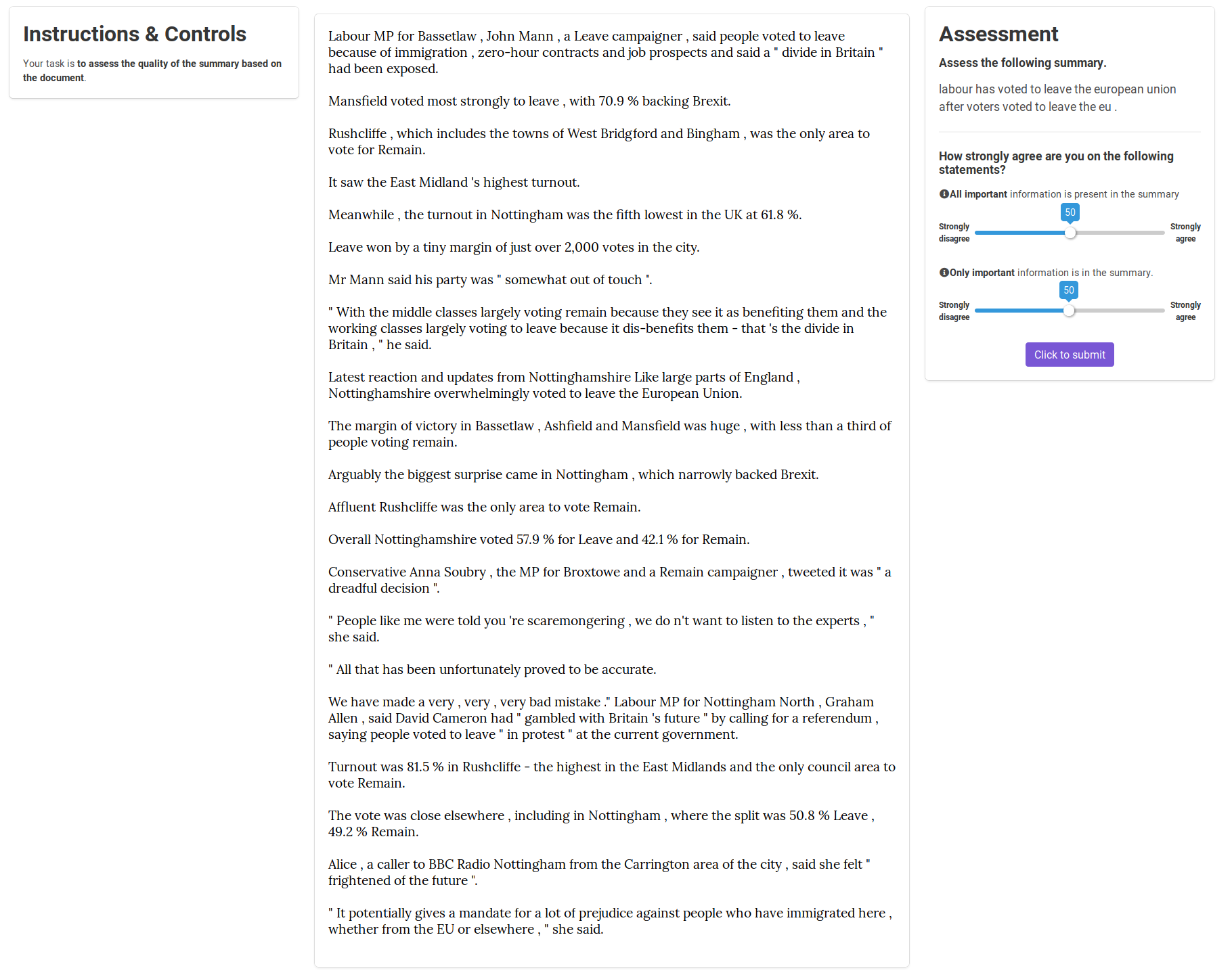}}
    \caption{The UI for content evaluation without highlight. At the right of the page judges give the recall and precision assessment by sliding the scroller from 1 to 100 based on the given summary quality.}
    \label{fig:my_label}
\end{figure}
\begin{figure}[ht!]
    \centering
    \fbox{\includegraphics[width=17cm]{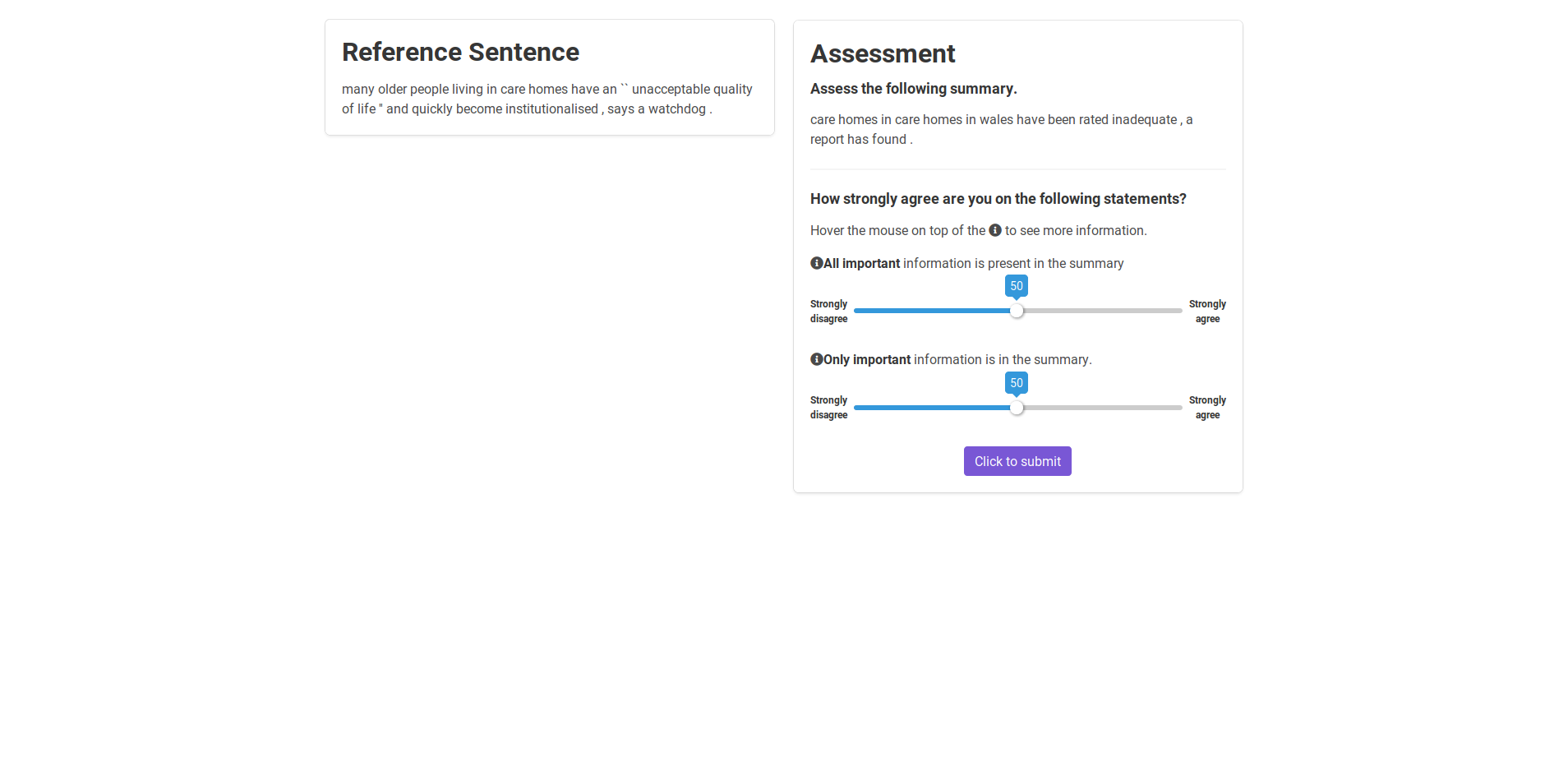}}
    \caption{The UI for content evaluation using reference summary as comparison. At the right of the page judges give the recall and precision assessment by sliding the scroller from 1 to 100 based on the given summary quality.}
\end{figure}
\newpage
\begin{figure}[ht!]
    \centering
    \fbox{\includegraphics[width=17cm]{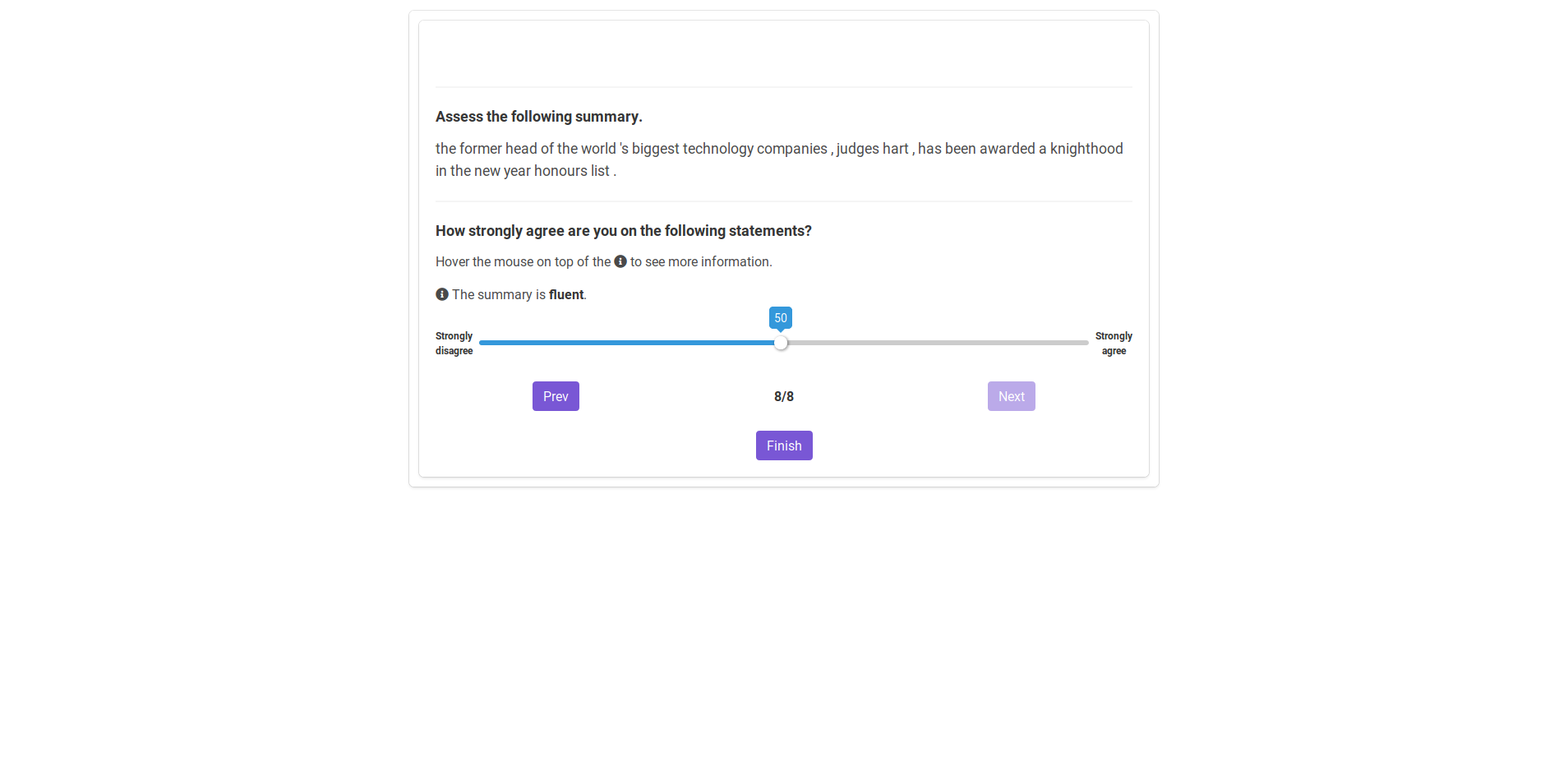}}
    \caption{The UI for fluency evaluation. Judges are given a number of summaries which can be switched by pressing the `Prev' or `Next' button. To give assessment, there is a scroller from 1 to 100.}
\end{figure}
\begin{figure}[ht!]
    \centering
    \includegraphics[width=17cm]{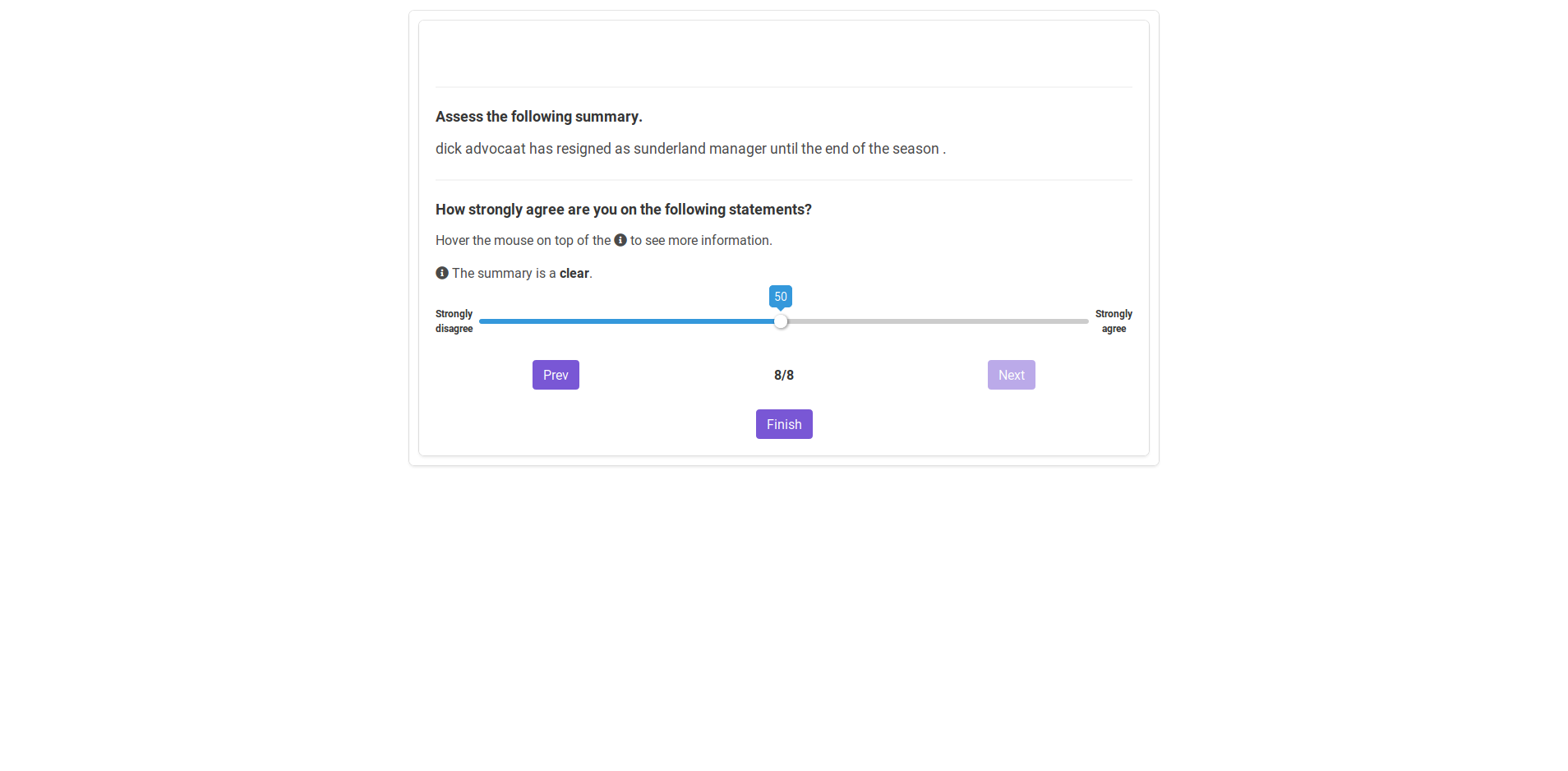}
    \caption{The UI for clarity evaluation. Judges are given a number of summaries which can be switched by pressing the `Prev' or `Next' button. To give assessment, there is a scroller from 1 to 100.}
\end{figure}
\end{document}